\documentclass[11pt,a4paper]{article}
\usepackage[hyperref]{conll-2019}
\usepackage{times}
\usepackage{latexsym}
\usepackage{times}  
\usepackage{helvet}  
\usepackage{courier}  
\usepackage{url}  
\usepackage{graphicx}  
\usepackage{amsfonts}
\usepackage{amsmath}
\usepackage{booktabs}
\usepackage{comment}
\usepackage[font=small]{caption}
\usepackage{multirow}
\aclfinalcopy 

\title{Learning to Represent Bilingual Dictionaries}

\author{Muhao Chen$^{1}$\thanks{\indent Both authors contributed equally to this work.} , Yingtao Tian$^{2*}$, Haochen Chen$^2$,\\ \textbf{Kai-Wei Chang}$^1$, \textbf{Steven Skiena}$^2$ \& \textbf{Carlo Zaniolo}$^1$\\
  $^1$University of California, Los Angeles, CA, USA \\
  $^2$The State University of New York, Stony Brook, NY, USA \\
  \texttt{muhaochen@ucla.edu}; \texttt{\{kwchang, zaniolo\}@cs.ucla.edu};\\ \texttt{\{yittian, haocchen, skiena\}@cs.stonybrook.edu} }

\date{}

\begin{document}
\maketitle

\begin{abstract}
Bilingual word embeddings have been widely used to capture the correspondence of lexical semantics in different human languages.
However, the cross-lingual correspondence between sentences and words is less studied, despite that this correspondence can significantly benefit many applications
such as cross-lingual semantic search and textual inference.
To bridge this gap, we propose a neural embedding model that leverages bilingual dictionaries\footnote{We refer the term \emph{dictionary} to its common meaning, i.e. lexical definitions of words. Note that this is different from some papers on bilingual settings that refer dictionaries to seed lexicons for one-to-one word mappings.}.
The proposed model is trained to map the lexical definitions to the cross-lingual target words,
for which we explore with different sentence encoding techniques.
To enhance the learning process on limited resources, our model adopts several critical learning strategies, including multi-task learning on different bridges of languages, and joint learning of the dictionary model with a bilingual word embedding model.
We conduct experiments on two new tasks.
In the cross-lingual reverse dictionary retrieval task, we demonstrate that our model is capable of comprehending bilingual concepts based on descriptions, and the proposed learning strategies are effective.
In the bilingual paraphrase identification task, we show that our model effectively associates sentences in different languages via a shared embedding space, and outperforms existing approaches in identifying bilingual paraphrases.
\end{abstract}

\newcommand\BibTeX{B\textsc{ib}\TeX}
\newcommand{\inv}{\vspace{-1em}}
\newcommand{\stitle}[1]{\vspace{0.0ex}\noindent{\bf #1}}
\newcommand{\modelnamens}{\mbox{\texttt{BilDRL}}}
\newcommand{\modelname}{{\modelnamens}\ }

\section{Introduction}

Cross-lingual semantic representation learning has attracted significant attention recently.
Various approaches have been proposed to align words of different languages in a shared embedding space \cite{ruder2017survey}.
By offering task-invariant semantic transfers, these approaches critically support many cross-lingual NLP tasks including neural machine translations (NMT) \cite{devlin2014fast},
bilingual document classification \cite{zhou2016cross}, knowledge alignment \cite{chen2018cotrain} and entity linking \cite{upadhyay2018joint}.
\par

While many existing approaches have been proposed to associate lexical semantics between languages \cite{ap2014autoencoder,gouws2015bilbowa,luong2015bilingual},
modeling the correspondence between lexical and sentential semantics across different languages is still
an unresolved challenge.
We argue that learning to represent such cross-lingual and multi-granular correspondence is well desired and natural for multiple reasons.
One reason is that, learning word-to-word correspondence has a natural limitation,
considering that many words do not have direct translations in another language.
For example, \emph{schadenfreude} in German, which means \emph{a feeling of joy that comes from knowing the troubles of other people}, has no proper English counterpart word.
To appropriately learn the representations of such words in bilingual embeddings,
we need to capture their meanings based on the definitions.\par

\begin{figure}
  \centering
  \includegraphics[width=0.96\columnwidth]{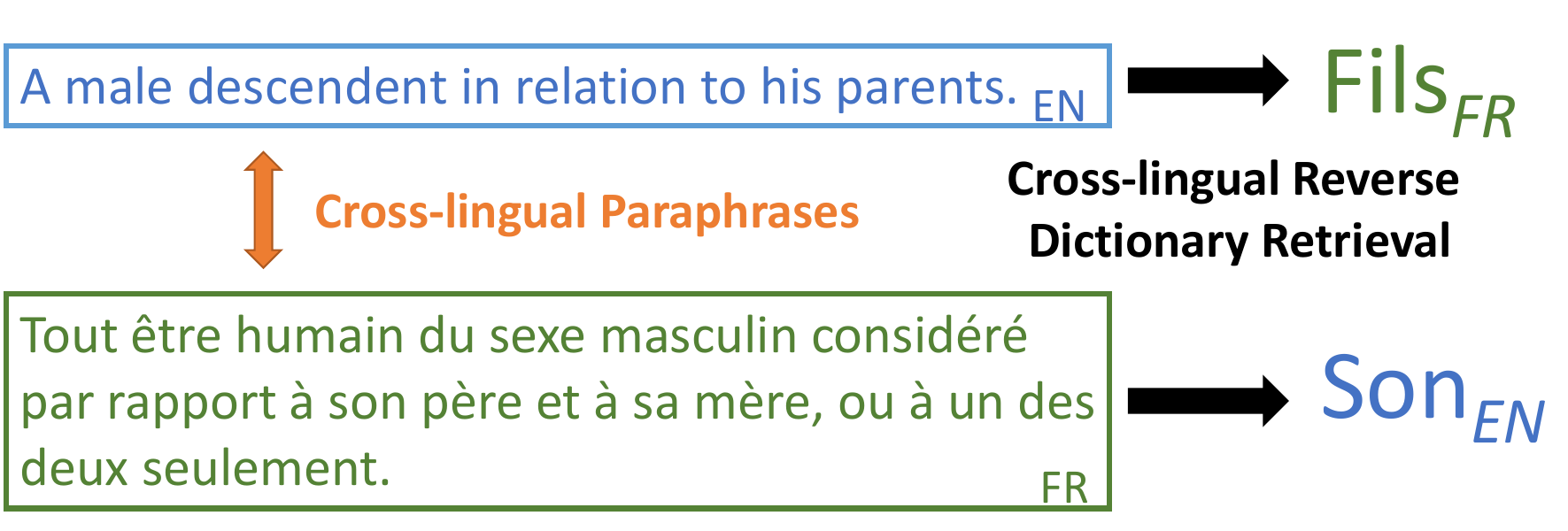}
  \caption{An example illustrating the two cross-lingual tasks.
  The \emph{cross-lingual reverse dictionary retrieval} finds cross-lingual target words based on descriptions.
  In terms of \emph{cross-lingual paraphrases}, the French sentence
  (which means \emph{any male being considered in relation to his father and mother, or only one of them})
  describes the same meaning as the English sentence, but has much more content details.}~\label{fig:intro-fig}
\end{figure}

Besides, modeling such correspondence is also highly beneficial to many application scenarios. 
One example is cross-lingual semantic search of concepts \cite{hill2016learning}, where the lexemes or concepts are retrieved based on sentential descriptions (see Fig.~\ref{fig:intro-fig}).
Others include
discourse relation detection in bilingual dialogue utterances \cite{jiang2018learning}, multilingual text summarization \cite{nenkova2012survey},
and educational applications for foreign language learners.
Finally, it is natural in foreign language learning that a human learns foreign words by looking up their meanings in the native language \cite{hulstijn1996incidental}. Therefore, learning such correspondence essentially mimics human learning behaviors.
\par
However, realizing such a representation learning model is a non-trivial task, inasmuch as it requires a comprehensive learning process to effectively compose the semantics of arbitrary-length sentences in one language, and associate that with single words in another language.
Consequently, this objective also demands high-quality cross-lingual alignment that bridges between single and sequences of words.
Such alignment information is generally not available in the parallel and seed-lexicon that are utilized by bilingual word embeddings \cite{ruder2017survey}.\par

To incorporate the representations of bilingual lexical and sentential semantics, 
we propose an approach 
to capture \emph{the mapping from the definitions to the corresponding foreign words} by leveraging \emph{bilingual dictionaries}
The proposed model \mbox{\modelname} (\textbf{Bil}ingual \textbf{D}ictionary \textbf{R}epresentation \textbf{L}earning) first constructs a word embedding space 
with 
pre-trained bilingual word embeddings.
Based on cross-lingual word definitions, a sentence encoder is trained to realize the mapping from literal descriptions to target words in the bilingual word embedding space,
for which 
we investigate with multiple 
encoding techniques.
To enhance 
cross-lingual learning 
on limited resources, \modelname conducts multi-task learning on different directions of a language pair.
Moreover, \modelname enforces a joint learning strategy of bilingual word embeddings and the sentence encoder, which seeks to gradually adjust the embedding space to better suit the representation of cross-lingual word definitions.\par

To show the applicability of \modelnamens, we conduct experiments on two useful cross-lingual tasks (see Fig.~\ref{fig:intro-fig}). 
(i) \emph{Cross-lingual reverse dictionary retrieval} seeks to retrieve words or concepts 
given descriptions in another language.
This task is useful to help users find foreign words 
based on the notions or descriptions,
and is especially beneficial to users such as translators, foreigner language learners and technical writers using non-native languages.
We show that \mbox{\modelname} achieves promising results on this task, while bilingual multi-task learning and joint learning dramatically enhance the performance.
(ii) \emph{Bilingual paraphrase identification} asks whether two sentences in different languages essentially express the same meaning, which is critical to question answering or dialogue systems that apprehend multilingual utterances \cite{bannard2005paraphrasing}.
This task is challenging, as it requires a model to comprehend cross-lingual paraphrases that are inconsistent in grammar, content details and word orders. 
\modelname maps sentences to the lexicon embedding space. 
This process reduces the problem to evaluate the similarity of lexicon embeddings, which can be easily solved by a simple classifier. 
\modelname performs well with even a small amount of data, and significantly outperforms previous approaches. 


\section{Related Work}\label{sec:rel}
We discuss two lines of relevant work.

\stitle{Bilingual word embeddings}.
Various approaches have been proposed for training bilingual word embeddings. These approaches span in two families: off-line mappings and joint training. \par
The off-line mapping based approach fixes the structures of pre-trained monolingual embeddings, and induces bilingual projections based on seed lexicons \cite{mikolov2013exploiting}.
Some variants of this approach improve the quality of projections by adding constraints such as orthogonality of transforms, normalization and mean centering of embeddings \cite{xing2015normalized,artetxe2016learning,vulic2016role}.
Others adopt canonical correlation analysis to map separate monolingual embeddings to a shared embedding space \cite{faruqui2014improving,doval2018improving}.\par

Unlike off-line mappings, joint training models simultaneously update word embeddings and cross-lingual alignment.
In doing so, such approaches generally capture more precise cross-lingual semantic transfer \cite{ruder2017survey,upadhyay2018joint}.
While a few such models still maintain separated embedding spaces for each language \cite{artetxe2017learning},
more of them maintain a unified space for both languages.
The cross-lingual semantic transfer by these models is captured from parallel corpora with sentential or document-level alignment,
using techniques such as
bilingual bag-of-words distances (BilBOWA) \cite{gouws2015bilbowa}, Skip-Gram \cite{coulmance2015trans} and sparse 
tensor factorization \cite{vyas2016sparse}.\par

\stitle{Neural sentence modeling}. Neural sentence models seek to capture phrasal or sentential semantics from word sequences.
They often adopt encoding techniques such as recurrent neural encoders (RNN) \cite{kiros2015skip}, convolutional encoders (CNN) \cite{chen2018neural}, and attentive encoders \cite{rocktaschel2015reasoning} to represent the composed semantics of a sentence as an embedding vector.
Recent works have focused on apprehending pairwise correspondence of sentential semantics by adopting multiple neural sentence models in one learning architecture,
including Siamese models for detecting discourse relations of sentences \cite{sha2016reading}, and sequence-to-sequence models for tasks like style transfer \cite{shen2017style}, text summarization \cite{chopra2016abstractive} and translation \cite{wu2016google}.

\par
On the other hand, fewer efforts have been put to characterizing the associations between sentential and lexical semantics.
\citeauthor{hill2016learning} \shortcite{hill2016learning} and \citeauthor{xie2016representation} \shortcite{xie2016representation} learn off-line mappings between monolingual descriptions and lexemes to capture such associations. 
\citeauthor{eisner2016emoji2vec} \shortcite{eisner2016emoji2vec} adopt a similar approach to capture emojis based on descriptions.
At the best of our knowledge, there has been no previous approach to learn to discover the correspondence of sentential and lexical semantics in a multilingual scenario.
This is exactly the focus of our work, in which the proposed strategies of multi-task learning and joint learning are critical to the corresponding 
learning process under limited resources.
Utilizing such correspondence, our approach also sheds light on addressing discourse relation detection in a multilingual scenario.

\begin{figure*}
  \centering
  \includegraphics[width=0.99\textwidth]{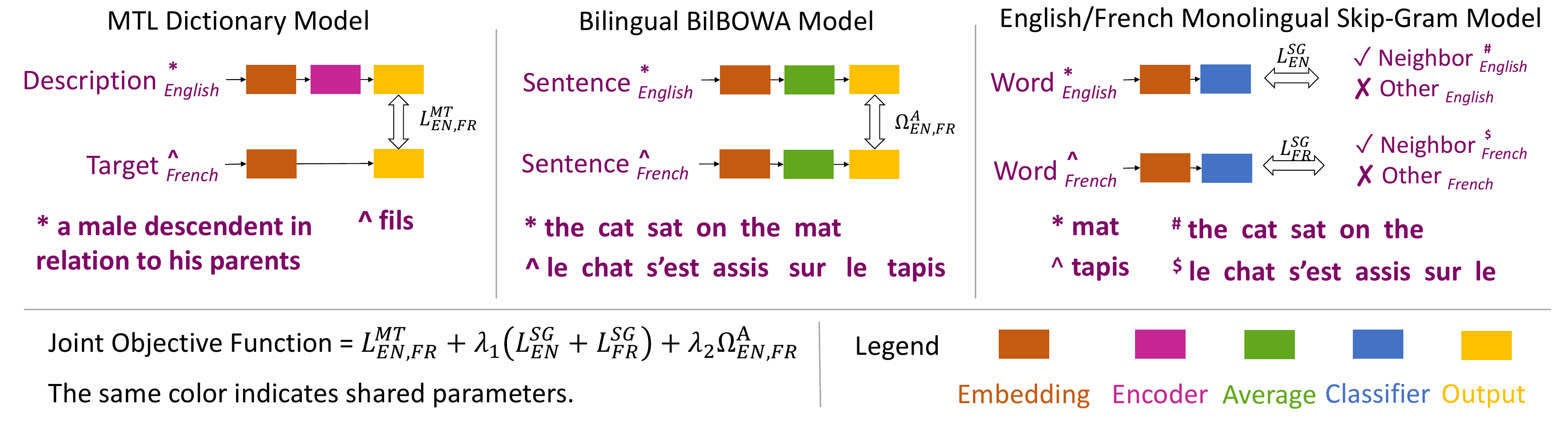}
  \caption{Joint learning architecture of \modelnamens.}~\label{fig:arch-fig}
\end{figure*} 
\def\lang{\mathcal{L}}
\def\alang{l}
\def\bhline{\specialrule{.2em}{0em}{0em}}
\section{Modeling Bilingual Dictionaries}\label{sec:model}
We hereby begin our modeling with the formalization of bilingual dictionaries.
We use $\lang$ to denote the set of languages. 
For a language $\alang \in \lang$, $V_{\alang}$ denotes its vocabulary, where for each word $w \in V_{\alang}$, bold-faced $\mathbf{w}\in \mathbb{R}^k$ denotes its embedding vector.
A $\alang_i$-$\alang_j$ bilingual dictionary $D(\alang_i, \alang_j)$ (or simply $D_{ij}$) contains dictionary entries $(w^i, S_{w}^j) \in D_{ij}$, in which $w^i \in V_{\alang_i}$, and 
$S_w^j=w_1^j\ldots w_n^j \; (w_\cdot^j \in V_{\alang_j})$
is a cross-lingual definition that describes the word $w^i$ with a sequence of words in language $\alang_j$.
For example, a French-English dictionary $D(\mathrm{Fr}, \mathrm{En})$ could include a French word \emph{app\'etite} accompanied by its English definition \emph{desire for, or relish of food or drink}.
Note that, for a word $w^i$, multiple definitions in $\alang_j$ may coexist.\par
\modelname is constructed and improved through three stages, as depicted in Fig.~\ref{fig:arch-fig}.
A sentence encoder is first used to learn from a bilingual dictionary the association between words and definitions.
Then in a pre-trained bilingual word embedding space, multi-task learning is conducted on both directions of a language pair.
Lastly, joint learning with word embeddings is enforced to simultaneously adjust the embedding space during the training of the dictionary model, which further enhances the cross-lingual learning process.

It is noteworthy that, NMT \cite{wu2016google} is considered as an ostensibly relevant method to ours. 
NMT does not apply to our problem setting bacause it has major differences from our work in those perspectives:
(i)
In terms of data modalities, NMT has to bridge between corpora of the same granularity, i.e. either between sentences or between lexicons. 
This is unlike \modelname that captures multi-granular correspondence of semantics across different modalities, i.e. sentences and words; 
(ii) As for learning strategies, NMT relies on an encoder-decoder architecture using end-to-end training \cite{luong2015effective}, while \modelname employs joint learning of a dictionary-based sentence encoder and a bilingual embedding space.

\subsection{Encoders for Lexical Definitions}
\modelname models a dictionary using a neural sentence encoder $E(S)$, which composes the meaning of the sentence into a latent vector representation.
We hereby introduce this model component, which is designed to be a GRU encoder with self-attention.
Besides that, we also investigate other widely-used neural sequence encoders.

\subsubsection{Attentive GRU Encoder}
\stitle{The GRU encoder} is a computationally efficient alternative of the LSTM~\cite{cho2014learning}. 
Each unit consists of a reset gate $\mathbf{r}_t$ and an update gate $\mathbf{z}_t$ to track the state of the sequence.
Given the vector representation $\mathbf{w}_t$ of an incoming item $w_t$, GRU updates the hidden state $\mathbf{h}^{(1)}_t$
as a linear combination of the previous state $\mathbf{h}^{(1)}_{t-1}$ and the candidate state $\tilde{\mathbf{h}}^{(1)}_t$ of the new item $w_t$ as below.

\inv
\begin{equation*}
\mathbf{h}^{(1)}_t=\mathbf{z}_t\odot \tilde{\mathbf{h}}^{(1)}_t+(1-\mathbf{z}_t)\odot \mathbf{h}^{(1)}_{t-1}
\end{equation*}
\inv

The update gate $\mathbf{z}_t$ balances between the information of the previous sequence and the new item,
where $\mathbf{M}_z$ and $\mathbf{N}_z$ are two weight matrices, $\mathbf{b}_z$ is a bias vector, and $\sigma$ is the sigmoid function.

\inv
\begin{equation*}
\mathbf{z}_t=\sigma\left (\mathbf{M}_z\mathbf{x}_t+\mathbf{N}_z \mathbf{h}^{(1)}_{t-1} + \mathbf{b}_z\right )
\end{equation*}
\inv

The candidate state $\tilde{\mathbf{h}}^{(1)}_t$ is calculated similarly to those in a traditional recurrent unit as below.
The reset gate $\mathbf{r}_t$ thereof controls how much information of the past sequence should contribute to the candidate state.

\inv
\begin{align*}
&\tilde{\mathbf{h}}^{(1)}_t=\mathrm{tanh}\left (\mathbf{M}_s\mathbf{w}_t+\mathbf{r}_t\odot(\mathbf{N}_s \mathbf{h}^{(1)}_{t-1}) + \mathbf{b}_s\right )\\
&\mathbf{r}_t=\sigma\left (\mathbf{M}_r\mathbf{w}_t+\mathbf{N}_r \mathbf{h}^{(1)}_{t-1} + \mathbf{b}_r\right )
\end{align*}
\inv

While a GRU encoder can 
stack multiple of the above GRU layers, without an attention mechanism, the last state $\mathbf{h}^{(1)}_S$ of the last layer represents the overall meaning
of the encoded sentence $S$.

\stitle{The self-attention mechanism} \cite{conneau2017supervised} seeks to 
highlight the important units in an input sentence when capturing its overall meaning, which is calculated as below:

\inv
\begin{align*}
\label{eq:attention}
&\mathbf{u}_t=\mathrm{tanh}\left (\mathbf{M}_a\mathbf{h}^{(1)}_t+\mathbf{b}_a\right )\\
&a_t=\frac{\mathrm{exp}\left (\mathbf{u}^\top_t\mathbf{u}_S\right )}{\sum_{w_m\in S}\mathrm{exp}\left (\mathbf{u}^\top_m\mathbf{u}_S\right )}\\
&\mathbf{h}^{(2)}_t=|S|a_t\mathbf{u}_t
\end{align*}
\inv

\noindent$\mathbf{u}_t$ is the intermediary representation of GRU output $\mathbf{h}^{(1)}_t$, and
$\mathbf{u}_S=\mathrm{tanh} (\mathbf{M}_a\mathbf{h}^{(1)}_S+\mathbf{b}_a )$ is 
that of the last GRU output $\mathbf{h}^{(1)}_S$. $\mathbf{u}_S$ can be seen as a high-level representation of the input sequence.
By measuring the similarity of each $\mathbf{u}_t$ with $\mathbf{u}_S$, the normalized attention weight $a_t$, which highlights an input that contributes significantly to the overall meaning, is produced through a softmax.
Note that a scalar $|S|$ is multiplied along with $a_t$ to $\mathbf{u}_t$, 
so as to keep the weighted representation $\mathbf{h}^{(2)}_t$
from losing the scale of $\mathbf{h}^{(1)}_t$.
The sentence encoding is calculated as the average of the last attention layer 
$E^{(1)}(S)=\frac{1}{|S|}\sum_{t=1}^{|S|}a_t\mathbf{h}_t^{(2)}$.\par

\subsubsection{Other Encoders}
We also experiment with other widely used neural sentence modeling techniques\footnote{Note that recent advances in monolingual contextualized embeddings like multilingual ELMo~\cite{peters2018deep,che2018conll} and M-BERT~\cite{pires2019multilingual,devlin2018bert} can also be supported to represent sentences for our setting. 
We leave them as future work, as they require non-trivial adaption to both multilingual settings and joint training, and extensive pre-training on external corpora.}, which are however outperformed by the attentive GRU in our tasks.
These techniques include the vanilla GRU, CNN \cite{kalchbrenner2014convolutional}, and linear bag-of-words (BOW) \cite{hill2016learning}.
We briefly introduce the later two techniques in the following.\par

\stitle{A convolutional encoder} 
applies a kernel $\mathbf{M}_c\in \mathbb{R}^{h \times k}$ to produce the latent representation $\mathbf{h}^{(3)}_t=\mathrm{tanh}(\mathbf{M}_c \mathbf{w}_{t:t+h-1} + \mathbf{b}_c)$ from each $h$-gram of the input vector sequence $\mathbf{w}_{t:t+h-1}$,
for which $h$ is the kernel size and $\mathbf{b}_c$ is a bias vector.
A sequence of latent vectors $\mathbf{H}^{(3)}=[ \mathbf{h}^{(3)}_1, \mathbf{h}^{(3)}_2, ..., \mathbf{h}^{(3)}_{|S|-h+1}]$ is produced from the input,
where each latent vector leverages the significant local semantic features from each $h$-gram.
Following convention~\cite{liu2017deep}, we apply dynamic max-pooling to extract robust features from the convolution outputs, and use the mean-pooling results of the last layer to represent the sentential semantics.

\stitle{The Linear bag-of-words (BOW) encoder} \cite{xie2016representation,hill2016learning} is realized by the sum of projected word embeddings of the input sentence, i.e. $E^{(2)}(S)=\sum_{t=1}^{|S|}\mathbf{M}_b\mathbf{w}_t$.

\subsection{Basic Learning Objective}\label{sec:dict-learning-objective}
The objective of learning the dictionary model is to map the encodings of cross-lingual word definitions to the target word embeddings.
This is realized by minimizing the following $L_2$ loss,

\inv
\begin{equation*}
  L_{ij}^\mathrm{ST}=\frac{1}{|D_{ij}|}\sum_{(w^i,S_w^j)\in D_{ij}}\left \| E_{ij}(S_w^j)-\mathbf{w}^i \right \|_2^2
\end{equation*}
\inv

\noindent
in which $E_{ij}$ is the dictionary model that maps from descriptions in $\alang_j$ to words in $\alang_i$.\par
The above defines the basic model variants of \modelname that learns on a single dictionary.
For word representations in the learning process, \mbox{\modelname} initializes the embedding space using pre-trained word embeddings.
Note that, without adopting the joint learning strategy in Section~\ref{sec:jo}, the learning process does not update word embeddings that are used to represent the definitions and target words.
While other forms of loss such as cosine proximity \cite{hill2016learning} and hinge loss \cite{xie2016representation} may also be used in the learning process,
we find that $L_2$ loss consistently leads to better performance in our experiments.

\subsection{Bilingual Multi-task Learning}\label{sec:mtl}
In cases where entries in a bilingual dictionary are not amply provided,
learning the above bilingual dictionary on one ordered language pair may fall short in insufficiency of alignment information.
One practical solution is to conduct a bilingual multi-task learning process.
In detail, given a language pair $(\alang_i, \alang_j)$, we learn the dictionary model $E_{ij}$ on both dictionaries $D_{ij}$ and $D_{ji}$ with shared parameters. Correspondingly, we rewrite the previous learning objective function as below, in which $D=D_{ij}\cup D_{ji}$.

\inv
\begin{equation*}
  L_{ij}^\mathrm{MT}=\frac{1}{|D|}\sum_{(w,S_w)\in D}\left \| E_{ij}(S_w)-\mathbf{w} \right \|_2^2
\end{equation*}
\inv

This strategy non-trivially requests the same dictionary model to represent semantic transfer in two directions of the language pair.
To fulfill such a request, we initialize the embedding space using the BilBOWA embeddings \cite{gouws2015bilbowa}, which provide a unified embedding space that resolves both monolingual and cross-lingual semantic relatedness of words. 
In practice, we find this simple multi-task strategy to bring significant improvement to our cross-lingual tasks.
Note that, besides BilBOWA, other joint-training bilingual embeddings in a unified space \cite{doval2018improving} can also support this strategy,
for which we leave the comparison to future work.
\subsection{Joint Learning Objective}\label{sec:jo}
While above learning strategies are based on a fixed embedding space, we lastly propose a joint learning strategy. 
During the training process, this strategy simultaneously updates the embedding space based on both the dictionary model and the bilingual word embedding model.
The learning is through asynchronous minimization of the following joint objective function,

\inv
\begin{equation*}
  J=L_{ij}^\mathrm{MT}+\lambda_1 (L_{i}^\mathrm{SG}+L_{j}^\mathrm{SG})+\lambda_2\Omega_{ij}^\mathrm{A}
\end{equation*}
\inv

\noindent
where $\lambda_1$ and $\lambda_2$ are two positive hyperparameters.
$L_{i}^\mathrm{SG}$ and $L_{j}^\mathrm{SG}$ are the original Skip-Gram losses \cite{mikolov2013word2vec} 
to separately obtain word embeddings on monolingual corpora of ${\alang}_i$ and ${\alang}_j$.
$\Omega_{ij}^\mathrm{A}$, termed as below, is the alignment loss to minimize  bag-of-words distances for aligned sentence pairs $(S^i,S^j)$ in parallel corpora $C_{ij}$.

\inv
\begin{align*}
&\Omega_{ij}^\mathrm{A} = \frac{1}{|C_{ij}|}\sum_{(S^i, S^j)\in C_{ij}} d_{\mathrm{S}}(S^i,S^j)\\
&d_{\mathrm{S}}(S^i,S^j) = \left\| \frac{1}{|S^i|}\sum_{w^i_m\in S^i}\mathbf{w}_m^i  - \frac{1}{|S^j|}\sum_{w^j_n\in S^j}\mathbf{w}_n^j \right\|_2^2
\end{align*}
\inv


The joint learning process adapts the embedding space to better suit the dictionary model, which is shown to further enhance the cross-lingual learning of \modelnamens.
\subsection{Training}
To initialize the embedding space, we pre-trained BilBOWA
on the parallel corpora Europarl v7 \cite{koehn2005europarl} and monolingual corpora of tokenized Wikipedia dump \cite{al2013polyglot}.
For models without joint learning, we use \mbox{AMSGrad} \cite{reddi2018convergence} to optimize the parameters.
Each model without bilingual multi-task learning thereof, is trained on batched samples from each individual dictionary.
Multi-task learning models are trained on batched samples from two dictionaries.
Within each batch, entries of different directions of languages can be mixed together. 
For joint learning, we 
conduct an efficient 
multi-threaded asynchronous training \cite{mnih2016asynchronous} of \mbox{AMSGrad}.
In detail, after initializing the embedding space based on pre-trained BilBOWA, parameter updating based on the four components of $J$ occurs across four worker threads.
Two monolingual threads select batches of monolingual contexts from the Wikipedia dump of two languages for Skip-Gram,
one alignment thread randomly samples parallel sentences from Europarl,
and one dictionary thread extracts samples of entries for a bilingual multi-task dictionary model.
Each thread makes a batched update to model parameters asynchronously for each term of $J$.
The asynchronous training of all threads keeps going until the dictionary thread finishes its epochs. 
\def\hitsone{\mathit{P}\mbox{@}1}
\def\hitsten{\mathit{P}\mbox{@}10}
\def\hitsh{\mathit{P}\mbox{@}100}
\def\mean{\mathit{Mean}}
\def\mrr{\mathit{MRR}}

\section{Experiments}

We present 
experiments on two multilingual tasks: the cross-lingual reverse dictionary retrieval task and the bilingual paraphrase identification task.

\subsection{Datasets}

The experiment of cross-lingual reverse dictionary retrieval is conducted on a trilingual dataset \emph{Wikt3l}.
This dataset is extracted from Wiktionary\footnote{\url{https://www.wiktionary.org/}}, 
which is one of the largest freely available multilingual dictionary resources on the Web.
Wikt3l contains dictionary entries of language pairs (English, French) and (English, Spanish), which form En-Fr, Fr-En, En-Es and Es-En dictionaries on four bridges of languages in total.
Two types of cross-lingual definitions are extracted from Wiktionary: (i) cross-lingual definitions provided under the \emph{Translations} sections of Wiktionary pages; (ii) monolingual definitions for words that are linked to a cross-lingual counterpart with a inter-language link\footnote{An inter-language link matches the entries of counterpart words between language versions of Wiktionary. 
} of Wiktionary.
We exclude all the definitions of stop words in constructing the dataset, and list the statistics in Table~\ref{tbl:statistics}.\par

Since existing datasets for paraphrase identification are merely monolingual, 
we contribute with another dataset \emph{WBP3l} for cross-lingual sentential paraphrase identification.
This dataset contains 6,000 pairs of bilingual sentence pairs respectively for En-Fr and En-Es settings.
Within each bilingual setting, positive cases are formed as pairs of descriptions aligned by inter-language links, which exclude the word descriptions in Wikt3l for training \modelnamens.
To generate negative examples, given a source word, we first find its 15 nearest neighbors in the embedding space. 
Within the nearest neighbors, we use ConceptNet \cite{speer2017conceptnet} to filter out the synonyms of the source word, so as to prevent from generating false negative cases. 
Then we randomly pick one word from the filtered neighbors and pair its cross-lingual definition with the English definition of the source word to create a negative case.
This process ensures that each negative case is endowed with limited dissimilarity of sentence meanings, which makes the decision more challenging.
For each language setting, we randomly select 70\% for training, 5\% for validation, and the rest 25\% for testing.
Note that each language setting of this dataset thereof, matches with the quantity and partitioning of sentence pairs in the widely-used Microsoft Research Paraphrase Corpus benchmark for monolingual paraphrase identification~\cite{yin2015abcnn,das2009paraphrase}.
Several examples from the dataset are shown in Table~\ref{tbl:paraph_exp}.
The datasets and the processing scripts are available at \url{https://github.com/muhaochen/bilingual_dictionaries}.

{
\begin{table}[t]
\setlength\tabcolsep{1pt}
\centering
\small
\begin{tabular}{c|cccc}
\bhline
Dictionary&En-Fr&Fr-En&En-Es&Es-En\\
\hline
\#Target words&15,666&16,857&8,004&16,986\\
\#Definitions&50,412&58,808&20,930&56,610\\
\bhline
\end{tabular}
\caption{Statistics of the bilingual dictionary dataset Wikt3l.}\label{tbl:statistics}
\end{table}
} 
{
\begin{table}[t]
\centering
{%
\setlength\tabcolsep{1pt}
\small
\begin{tabular}{cl}
\bhline
\multicolumn{2}{c}{Positive Examples}\\
\hline
\footnotesize\textbf{En}:&\emph{Being remote in space.}\\
\textbf{Fr}:&\emph{Se trouvant \`{a} une grande distance.}\\
\hline
\footnotesize\textbf{En}:&\emph{The interdisciplinary science that applies theories and}\\
&\emph{methods of the physical sciences to questions of biology.}\\
\textbf{Es}:&\emph{Ciencia que emplea y desarrolla las teorias y m\'etodos de }\\
&\emph{la f\'isica en la investigaci\'on de los sistemas biológicos.}\\
\bhline
\multicolumn{2}{c}{Negative Examples}\\
\hline
\footnotesize\textbf{En}:&\emph{A person who secedes or supports secession from a }\\
&\emph{political union.}\\
\textbf{Fr}:&\emph{Contr\^{o}le politique exerc\'{e} par une grande puissance sur }\\
&\emph{une contrée inf\'{e}od\'{e}e.}\\
\hline
\footnotesize\textbf{En}:&\emph{The fear of closed, tight places.}\\
\textbf{Es}:&\emph{P\'{e}rdida o disminuci\'{o}n considerables de la memoria.}\\
\bhline
\end{tabular}
}
\caption{Examples of bilingual paraphrases from WBP3l.}\label{tbl:paraph_exp}
\end{table}
} 
{
\begin{table*}[t]
\centering
{%
\setlength\tabcolsep{2pt}
\small
\begin{tabular}{c|ccc|ccc|ccc|ccc}
\bhline
Languages&\multicolumn{3}{c|}{En-Fr}&\multicolumn{3}{c|}{Fr-En}&\multicolumn{3}{c|}{En-Es}&\multicolumn{3}{c}{Es-En}\\
\hline
Metric&$\hitsone$&$\hitsten$&$\mrr$&$\hitsone$&$\hitsten$&$\mrr$&$\hitsone$&$\hitsten$&$\mrr$&$\hitsone$&$\hitsten$&$\mrr$\\
\bhline
BOW&0.8&3.4&0.011&0.4&2.2&0.006&0.4&2.4&0.007&0.4&2.6&0.007\\
CNN&6.0&12.4&0.070&6.4&14.8&0.072&3.8&7.2&0.045&7.0&16.8&0.088\\
GRU&35.6&46.0&0.380&38.8&49.8&0.410&47.8&59.0&0.496&57.6&67.2&0.604\\
ATT&38.8&47.4&0.411&39.8&50.2&0.425&51.6&59.2&0.534&60.4&68.4&0.629\\
\hline
GRU-mono&21.8&33.2&0.242&27.8&37.0&0.297&34.4&41.2&0.358&36.8&47.2&0.392\\
ATT-mono&22.8&33.6&0.249&27.4&39.0&0.298&34.6&42.2&0.358&39.4&48.6&0.414\\
\hline
GRU-MTL&43.4&49.2&0.452&44.4&52.8&0.467&50.4&60.0&0.530&63.6&71.8&0.659\\
ATT-MTL&46.8&56.6&0.487&47.6&56.6&0.497&55.8&62.2&0.575&66.4&75.0&0.687\\
\hline
ATT-joint&$\mathbf{63.6}$&$\mathbf{69.4}$&$\mathbf{0.654}$&$\mathbf{68.2}$&$\mathbf{75.4}$&$\mathbf{0.706}$&$\mathbf{69.0}$&$\mathbf{72.8}$&$\mathbf{0.704}$&$\mathbf{78.6}$&$\mathbf{83.4}$&$\mathbf{0.803}$\\
\bhline
\end{tabular}
}
\caption{Cross-lingual reverse dictionary retrieval results by \modelname variants. We report $\hitsone$, $\hitsten$, and $\mrr$ on four groups of models: (i) basic dictionary models that adopt four different encoding techniques (BOW, CNN, GRU and ATT); (ii) models with the two best encoding techniques that enforce the monolingual retrieval approach by \citeauthor{hill2016learning} \shortcite{hill2016learning} (GRU-mono and ATT-mono); (iii) models adopting bilingual multi-task learning (GRU-MTL and ATT-MTL); (iv) joint learning that employs the best dictionary model of ATT-MTL (ATT-joint).}\label{tbl:reverse}
\end{table*}
}

\subsection{Cross-lingual Reverse Dictionary Retrieval}

The objective of this task is to enable cross-lingual semantic retrieval of words based on descriptions.
Besides comparing variants of \modelname that adopt different sentence encoders and learning strategies, we also compare with the monolingual retrieval approach proposed by \citeauthor{hill2016learning} \shortcite{hill2016learning}.
Instead of directly associating cross-lingual word definitions, this approach learns definition-to-word mappings in a monolingual scenario.
When it applies to the multilingual setting, given a lexical definition, it first retrieves the corresponding word in the source language. Then, it looks for semantically related words in the target language using bilingual word embeddings.
As discussed in Section~\ref{sec:model}, NMT does not apply to this task due that it cannot capture the multi-granular correspondence between a sentence and a word.

\stitle{Evaluation Protocol.}
Before training the models, we randomly select 500 word definitions from each dictionary respectively as test cases, and exclude these definitions 
from the training data.
Each of the basic \modelname variants are trained on one bilingual dictionary.
The monolingual retrieval models are trained to fit the target words in the original languages of the word definitions, which are also provided in Wiktionary.
\modelname variants with multi-task or joint learning use both dictionaries of the same language pair.
In the test phase, for each test case $(w^i, S_w^j) \in D_{ij}$, the prediction performs a kNN search from the definition encoding $E_{ij}(S_w^j)$, and record the rank of $w^i$ within the vocabulary of ${\alang}_i$.
We limit the vocabularies to all words that appear in the Wikt3l dataset, which involve around 45k English words, 44k French words and 36k Spanish words.
To prevent the surface information of the target word from appearing in the definition, we have also masked out any translation of the target word occurring in the definition using a wildcard token \texttt{<concept>}.
We aggregate three metrics on test cases: the accuracy $\hitsone$ (\%), the proportion of ranks no larger than 10 $\hitsten$ (\%), and mean reciprocal rank $\mrr$.\par
We pre-train BilBOWA based on the original configuration by \citeauthor{gouws2015bilbowa}~\shortcite{gouws2015bilbowa} and obtain 50-dimensional initialization of bilingual word embedding spaces respectively for the English-French and English-Spanish settings.
For CNN, GRU, and attentive GRU (ATT) encoders, we stack five of each corresponding encoding layers with hidden-sizes of 200, 
and two affine layers are applied to the final output for dimension reduction.
This encoder architecture consistently represents the best performance through our tuning.
Through comprehensive hyperparameter tuning, we fix the learning rate $\alpha$ to 0.0005, the exponential decay rates of AMSGrad $\beta_1$ and $\beta_2$ to 0.9 and 0.999, coefficients $\lambda_1$ and $\lambda_2$ to both 0.1, and batch size to 64.
Kernel-size and pooling-size are both set to 2 for CNN.
Word definitions are zero-padded (short ones) or truncated (long ones) to the sequence length of 15, since most definitions (over $92\%$) are within 15 words in the dataset.
Training is limited to 1,000 epochs for all models as well as the dictionary thread of asynchronous joint learning, in which all models are able to converge.\par

\stitle{Results.} Results are reported in Table~\ref{tbl:reverse} in four groups.
The first group compares four different encoding techniques for the basic dictionary models.
GRU thereof consistently outperforms CNN and BOW, since the latter two fail to capture the important sequential information for descriptions.
ATT that weighs among the hidden states has notable improvements over GRU.
While we equip the two better encoding techniques with the monolingual retrieval approach (GRU-mono and ATT-mono), we find that the way of learning the dictionary models towards monolingual targets and retrieving cross-lingual related words incurs more impreciseness to the task.
For models of the third group that conduct multi-task learning in two directions of a language pair, the results show significant enhancement of performance in both directions.
For the final group of results, we incorporate the best variant of multi-task models into the joint learning architecture, which leads to compelling improvement of the task on all settings.
This demonstrates that properly adapting the word embeddings in joint with the bilingual dictionary model efficaciously constructs the embedding space that suits better the representation of both bilingual lexical and sentential semantics.\par

In general, this experiment has identified the proper encoding techniques of the dictionary model.
The proposed strategies of multi-task and joint learning effectively contribute to the precise characterization of the cross-lingual correspondence of lexical and sentential semantics, which have led to very promising capability of cross-lingual reverse dictionary retrieval.

\subsection{Bilingual Paraphrase Identification}

The bilingual paraphrase identification problem\footnote{Paraphrases have similar meanings, but can largely differ in content details and word orders. Hence, they are essentially different from translations. We have found that even the well-recognized Google NMT frequently caused distortions to short sentence meanings, and led to results that were close to random guess by the baseline classifiers after translation.} is a binary classification task with the goal to decide whether two sentences in different languages express the same meanings.
\modelname provides an effective 
solution by transferring sentential meanings to word-level representations and learning a simple classifier.
We evaluate three variants of \modelname on this task using WBP3l: 
the multi-task \modelname with GRU encoders (\modelnamens-GRU-MTL),
the multi-task \modelname with attentive GRU encoders (\modelnamens-ATT-MTL),
and the joint learning \modelname with with attentive GRU encoders (\modelnamens-ATT-joint).
We compare against several baselines of neural sentence pair models that are proposed 
for monolingual paraphrase identification.
These models include 
Siamese structures of CNN (BiCNN)~\cite{yin2015convolutional}, RNN 
(BiLSTM)~\cite{mueller2016siamese}, attentive CNN (ABCNN)~\cite{yin2015abcnn}, attentive GRU (BiATT)~\cite{rocktaschel2015reasoning},
and BOW (BiBOW).
To support the reasoning of cross-lingual semantics, we provide the baselines with the same BilBOWA embeddings.
\par
\stitle{Evaluation protocol.}
\modelname transfers each sentence into a vector in the word embedding space.
Then, for each sentence pair in the train set, a Multi-layer Perceptron (MLP) with a binary softmax loss is trained on the subtraction of two vectors as a downstream classifier.
Baseline models are trained end-to-end, each of which directly uses a parallel pair of encoders with shared parameters and an MLP that is stacked to the subtraction of two sentence vectors.
Note that some works use concatenation~\cite{yin2015convolutional} or Manhattan distances~\cite{mueller2016siamese} of sentence vectors instead of their subtraction~\cite{jiang2018learning}, which we find to be less effective on small amount of data.\par
We apply the configurations of the sentence encoders from the last experiment to corresponding baselines,
so as to show the performance under controlled variables.
Training of a classifier is terminated by early-stopping based on the validation set. 
Following convention~\cite{hu2014convolutional,yin2015abcnn}, we report the accuracy and F1 scores.\par

{
\begin{table}[t]
\centering
{%
\setlength\tabcolsep{1pt}
\small
\begin{tabular}{c|cc|cc}
\bhline
Languages&\multicolumn{2}{c|}{En\&Fr}&\multicolumn{2}{c}{En\&Es}\\
\hline
Metrics&\emph{Acc.}&\emph{F1}&\emph{Acc.}&\emph{F1}\\
\bhline
BiBOW&54.93&0.622&56.27&0.623\\
BiCNN&54.33&0.625&53.80&0.611\\
ABCNN&56.73&0.644&58.83&0.655\\
BiLSTM&59.60&0.662&57.60&0.637\\
BiATT&61.47&0.699&61.27&0.689\\
\hline
\multicolumn{1}{l|}{\modelnamens-GRU-MTL}&64.80&0.732&63.33&0.722\\
\multicolumn{1}{l|}{\modelnamens-ATT-MTL}&65.27&0.735&66.07&0.735\\
\multicolumn{1}{l|}{\modelnamens-ATT-joint}&$\mathbf{68.53}$&$\mathbf{0.785}$&$\mathbf{67.13}$&$\mathbf{0.759}$\\
\bhline
\end{tabular}
}
\caption{Accuracy and F1-scores of bilingual paraphrase identification. For \modelnamens, the results by three model variants are reported: \modelnamens-GRU-MTL and \modelnamens-ATT-MTL are models with bilingual multi-task learning, and \modelnamens-ATT-joint is the best ATT-based dictionary model variant deployed with both multi-task and joint learning.}\label{tbl:paraph}
\end{table}
} 

\stitle{Results.}
This task is challenging due to the heterogeneity of cross-lingual paraphrases and limitedness of learning resources.
The results in Table~\ref{tbl:paraph} show that all the baselines, 
where BiATT consistently outperforms the others, merely reaches slightly over 60\% of accuracy on both En-Fr and En-Es settings.
We believe that it comes down to the fact that sentences of different languages are often drastically heterogenous in both lexical semantics and the sentence grammar that governs the composition of words.
Hence, it is not surprising that previous neural sentence pair models, which capture the semantic relation of bilingual sentences directly from all participating words, fall short at the multilingual task.
\modelnamens, however, effectively leverages the correspondence of lexical and sentential semantics to simplify the task to an easier entailment task in the lexicon space, for which the multi-task learning \modelnamens-ATT-MTL outperforms the best baseline respectively by 3.80\% and 4.80\% of accuracy in both language settings, while \modelnamens-ATT-joint, employing the joint learning, further improves the task by another satisfying 3.26\% and 1.06\% of accuracy.
Both also show notable increment in F1.

\section{Conclusion and Future Work}

In this paper, we propose a neural embedding model \modelname 
that captures the correspondence of cross-lingual lexical and sentential semantics. 
We experiment with multiple forms of neural models and identify the best technique. 
The two learning strategies, bilingual multi-task learning and joint learning, are effective at enhancing the cross-lingual learning with limited resources,
and also achieve promising performance on cross-lingual reverse dictionary retrieval and bilingual paraphrase identification tasks by associating lexical and sentential semantics.
An important direction of future work is to explore whether the word-sentence alignment can improve bilingual word embeddings.
Applying \modelname to bilingual question answering and semantic search systems is another important direction. 

\section{Acknowledgement}

We thank the anonymous reviewers for their insightful comments.
This work was supported in part by National Science Foundation
Grant IIS-1760523.

\bibliography{ref}

\begin{thebibliography}{51}
\expandafter\ifx\csname natexlab\endcsname\relax\def\natexlab#1{#1}\fi

\bibitem[{Al-Rfou et~al.(2013)Al-Rfou, Perozzi, and Skiena}]{al2013polyglot}
Rami Al-Rfou, Bryan Perozzi, and Steven Skiena. 2013.
\newblock Plyglot: Distributed word representations for multilingual nlp.
\newblock In \emph{The SIGNLL Conference on Computational Natural Language
  Learning}.

\bibitem[{Artetxe et~al.(2016)Artetxe, Labaka, and
  Agirre}]{artetxe2016learning}
Mikel Artetxe, Gorka Labaka, and Eneko Agirre. 2016.
\newblock Learning principled bilingual mappings of word embeddings while
  preserving monolingual invariance.
\newblock In \emph{Proceedings of the 2016 Conference on Empirical Methods in
  Natural Language Processing}, pages 2289--2294.

\bibitem[{Artetxe et~al.(2017)Artetxe, Labaka, and
  Agirre}]{artetxe2017learning}
Mikel Artetxe, Gorka Labaka, and Eneko Agirre. 2017.
\newblock Learning bilingual word embeddings with (almost) no bilingual data.
\newblock In \emph{Proceedings of the 55th Annual Meeting of the Association
  for Computational Linguistics (Volume 1: Long Papers)}, volume~1, pages
  451--462.

\bibitem[{Bannard and Callison-Burch(2005)}]{bannard2005paraphrasing}
Colin Bannard and Chris Callison-Burch. 2005.
\newblock Paraphrasing with bilingual parallel corpora.
\newblock In \emph{Proceedings of the 43rd Annual Meeting on Association for
  Computational Linguistics}, pages 597--604. Association for Computational
  Linguistics.

\bibitem[{Chandar et~al.(2014)Chandar, Lauly, Larochelle, Khapra, Ravindran,
  Raykar, and Saha}]{ap2014autoencoder}
Sarath Chandar, Stanislas Lauly, Hugo Larochelle, Mitesh Khapra, Balaraman
  Ravindran, Vikas~C Raykar, and Amrita Saha. 2014.
\newblock An autoencoder approach to learning bilingual word representations.
\newblock In \emph{Advances in Neural Information Processing Systems}, pages
  1853--1861.

\bibitem[{Che et~al.(2018)Che, Liu, Wang, Zheng, and Liu}]{che2018conll}
Wanxiang Che, Yijia Liu, Yuxuan Wang, Bo~Zheng, and Ting Liu. 2018.
\newblock Towards better {UD} parsing: Deep contextualized word embeddings,
  ensemble, and treebank concatenation.
\newblock In \emph{CoNLL Shared Task}, pages 55--64. ACL.

\bibitem[{Chen et~al.(2018{\natexlab{a}})Chen, Meng, Huang, and
  Zaniolo}]{chen2018neural}
Muhao Chen, Chang~Ping Meng, Gang Huang, and Carlo Zaniolo. 2018{\natexlab{a}}.
\newblock Neural article pair modeling for wikipedia sub-article machine.
\newblock In \emph{Proceedings of European Conference of Machine Learning}.

\bibitem[{Chen et~al.(2018{\natexlab{b}})Chen, Tian, Chang, Skiena, and
  Zaniolo}]{chen2018cotrain}
Muhao Chen, Yingtao Tian, Kai-Wei Chang, Steven Skiena, and Carlo Zaniolo.
  2018{\natexlab{b}}.
\newblock Co-training embeddings of knowledge graphs and entity descriptions
  for cross-lingual entity alignment.
\newblock In \emph{Proceedings of the 27th International Joint Conference on
  Artificial Intelligence}.

\bibitem[{Cho et~al.(2014)Cho, van Merrienboer, Gulcehre, Bahdanau, Bougares,
  Schwenk, and Bengio}]{cho2014learning}
Kyunghyun Cho, Bart van Merrienboer, Caglar Gulcehre, Dzmitry Bahdanau, Fethi
  Bougares, Holger Schwenk, and Yoshua Bengio. 2014.
\newblock Learning phrase representations using rnn encoder--decoder for
  statistical machine translation.
\newblock In \emph{Proceedings of the 2014 Conference on Empirical Methods in
  Natural Language Processing (EMNLP)}, pages 1724--1734.

\bibitem[{Chopra et~al.(2016)Chopra, Auli, and Rush}]{chopra2016abstractive}
Sumit Chopra, Michael Auli, and Alexander~M Rush. 2016.
\newblock Abstractive sentence summarization with attentive recurrent neural
  networks.
\newblock In \emph{Proceedings of the 2016 Conference of the North American
  Chapter of the Association for Computational Linguistics: Human Language
  Technologies}, pages 93--98.

\bibitem[{Conneau et~al.(2017)Conneau, Kiela, Schwenk, Barrault, and
  Bordes}]{conneau2017supervised}
Alexis Conneau, Douwe Kiela, Holger Schwenk, Lo{\"\i}c Barrault, and Antoine
  Bordes. 2017.
\newblock Supervised learning of universal sentence representations from
  natural language inference data.
\newblock In \emph{Proceedings of the 2017 Conference on Empirical Methods in
  Natural Language Processing}, pages 670--680.

\bibitem[{Coulmance et~al.(2015)Coulmance, Marty, Wenzek, and
  Benhalloum}]{coulmance2015trans}
Jocelyn Coulmance, Jean-Marc Marty, Guillaume Wenzek, and Amine Benhalloum.
  2015.
\newblock Trans-gram, fast cross-lingual word-embeddings.
\newblock In \emph{Proceedings of the 2015 Conference on Empirical Methods in
  Natural Language Processing}, pages 1109--1113.

\bibitem[{Das and Smith(2009)}]{das2009paraphrase}
Dipanjan Das and Noah~A Smith. 2009.
\newblock Paraphrase identification as probabilistic quasi-synchronous
  recognition.
\newblock In \emph{Proceedings of the Joint Conference of the 47th Annual
  Meeting of the ACL and the 4th International Joint Conference on Natural
  Language Processing}, pages 468--476. Association for Computational
  Linguistics.

\bibitem[{Devlin et~al.(2018)Devlin, Chang, Lee, and
  Toutanova}]{devlin2018bert}
Jacob Devlin, Ming-Wei Chang, Kenton Lee, and Kristina Toutanova. 2018.
\newblock Bert: Pre-training of deep bidirectional transformers for language
  understanding.
\newblock \emph{arXiv preprint arXiv:1810.04805}.

\bibitem[{Devlin et~al.(2014)Devlin, Zbib, Huang, Lamar, Schwartz, and
  Makhoul}]{devlin2014fast}
Jacob Devlin, Rabih Zbib, Zhongqiang Huang, Thomas Lamar, Richard Schwartz, and
  John Makhoul. 2014.
\newblock Fast and robust neural network joint models for statistical machine
  translation.
\newblock In \emph{Proceedings of the 52nd Annual Meeting of the Association
  for Computational Linguistics (Volume 1: Long Papers)}, volume~1, pages
  1370--1380.

\bibitem[{Doval et~al.(2018)Doval, Camacho-Collados, Anke, and
  Schockaert}]{doval2018improving}
Yerai Doval, Jose Camacho-Collados, Luis~Espinosa Anke, and Steven Schockaert.
  2018.
\newblock Improving cross-lingual word embeddings by meeting in the middle.
\newblock In \emph{Proceedings of the 2018 Conference on Empirical Methods in
  Natural Language Processing}, pages 294--304.

\bibitem[{Eisner et~al.(2016)Eisner, Rockt{\"a}schel, Augenstein,
  Bo{\v{s}}njak, and Riedel}]{eisner2016emoji2vec}
Ben Eisner, Tim Rockt{\"a}schel, Isabelle Augenstein, Matko Bo{\v{s}}njak, and
  Sebastian Riedel. 2016.
\newblock emoji2vec: Learning emoji representations from their description.
\newblock \emph{arXiv preprint arXiv:1609.08359}.

\bibitem[{Faruqui and Dyer(2014)}]{faruqui2014improving}
Manaal Faruqui and Chris Dyer. 2014.
\newblock Improving vector space word representations using multilingual
  correlation.
\newblock In \emph{Proceedings of the 14th Conference of the European Chapter
  of the Association for Computational Linguistics}, pages 462--471.

\bibitem[{Gouws et~al.(2015)Gouws, Bengio, and Corrado}]{gouws2015bilbowa}
Stephan Gouws, Yoshua Bengio, and Greg Corrado. 2015.
\newblock Bilbowa: Fast bilingual distributed representations without word
  alignments.
\newblock In \emph{Proceedings of the 32nd International Conference on Machine
  Learning}, pages 748--756.

\bibitem[{Hill et~al.(2016)Hill, Cho, Korhonen, and Bengio}]{hill2016learning}
Felix Hill, KyungHyun Cho, Anna Korhonen, and Yoshua Bengio. 2016.
\newblock Learning to understand phrases by embedding the dictionary.
\newblock \emph{Transactions of the Association for Computational Linguistics},
  4:17--30.

\bibitem[{Hu et~al.(2014)Hu, Lu, Li, and Chen}]{hu2014convolutional}
Baotian Hu, Zhengdong Lu, Hang Li, and Qingcai Chen. 2014.
\newblock Convolutional neural network architectures for matching natural
  language sentences.
\newblock In \emph{Advances in neural information processing systems}, pages
  2042--2050.

\bibitem[{Hulstijn et~al.(1996)Hulstijn, Hollander, and
  Greidanus}]{hulstijn1996incidental}
Jan~H Hulstijn, Merel Hollander, and Tine Greidanus. 1996.
\newblock Incidental vocabulary learning by advanced foreign language students:
  The influence of marginal glosses, dictionary use, and reoccurrence of
  unknown words.
\newblock \emph{The modern language journal}, 80(3):327--339.

\bibitem[{Ji et~al.(2017)Ji, Liu, He, Zhao et~al.}]{xie2016representation}
Guoliang Ji, Kang Liu, Shizhu He, Jun Zhao, et~al. 2017.
\newblock Distant supervision for relation extraction with sentence-level
  attention and entity descriptions.
\newblock In \emph{Proceedings of the AAAI International Conference on
  Artificial Intelligence}, pages 3060--3066.

\bibitem[{Jiang et~al.(2018)Jiang, Chen, Chen, and Wang}]{jiang2018learning}
Jyun-Yu Jiang, Francine Chen, Yan-Ying Chen, and Wei Wang. 2018.
\newblock Learning to disentangle interleaved conversational threads with a
  siamese hierarchical network and similarity ranking.
\newblock In \emph{Proceedings of the 2018 Conference of the North American
  Chapter of the Association for Computational Linguistics: Human Language
  Technologies, Volume 1 (Long Papers)}, volume~1, pages 1812--1822.

\bibitem[{Kalchbrenner et~al.(2014)Kalchbrenner, Grefenstette, Blunsom,
  Kartsaklis, Kalchbrenner, Sadrzadeh, Kalchbrenner, Blunsom, Kalchbrenner, and
  Blunsom}]{kalchbrenner2014convolutional}
Nal Kalchbrenner, Edward Grefenstette, Phil Blunsom, Dimitri Kartsaklis, Nal
  Kalchbrenner, Mehrnoosh Sadrzadeh, Nal Kalchbrenner, Phil Blunsom, Nal
  Kalchbrenner, and Phil Blunsom. 2014.
\newblock A convolutional neural network for modelling sentences.
\newblock In \emph{Proceedings of the 52nd Annual Meeting of the Association
  for Computational Linguistics}, pages 212--217. Association for Computational
  Linguistics.

\bibitem[{Kiros et~al.(2015)Kiros, Zhu, Salakhutdinov, Zemel, Urtasun,
  Torralba, and Fidler}]{kiros2015skip}
Ryan Kiros, Yukun Zhu, Ruslan~R Salakhutdinov, Richard Zemel, Raquel Urtasun,
  Antonio Torralba, and Sanja Fidler. 2015.
\newblock Skip-thought vectors.
\newblock In \emph{Advances in neural information processing systems}, pages
  3294--3302.

\bibitem[{Koehn(2005)}]{koehn2005europarl}
Philipp Koehn. 2005.
\newblock Europarl: A parallel corpus for statistical machine translation.
\newblock In \emph{MT summit}, volume~5, pages 79--86.

\bibitem[{Liu et~al.(2017)Liu, Chang, Wu, and Yang}]{liu2017deep}
Jingzhou Liu, Wei-Cheng Chang, Yuexin Wu, and Yiming Yang. 2017.
\newblock Deep learning for extreme multi-label text classification.
\newblock In \emph{Proceedings of the 40th International ACM SIGIR Conference
  on Research and Development in Information Retrieval}, pages 115--124. ACM.

\bibitem[{Luong et~al.(2015{\natexlab{a}})Luong, Pham, and
  Manning}]{luong2015bilingual}
Thang Luong, Hieu Pham, and Christopher~D Manning. 2015{\natexlab{a}}.
\newblock Bilingual word representations with monolingual quality in mind.
\newblock In \emph{Proceedings of the 1st Workshop on Vector Space Modeling for
  Natural Language Processing}, pages 151--159.

\bibitem[{Luong et~al.(2015{\natexlab{b}})Luong, Pham, and
  Manning}]{luong2015effective}
Thang Luong, Hieu Pham, and Christopher~D Manning. 2015{\natexlab{b}}.
\newblock Effective approaches to attention-based neural machine translation.
\newblock In \emph{Proceedings of the 2015 Conference on Empirical Methods in
  Natural Language Processing}, pages 1412--1421.

\bibitem[{Mikolov et~al.(2013{\natexlab{a}})Mikolov, Le, and
  Sutskever}]{mikolov2013exploiting}
Tomas Mikolov, Quoc~V Le, and Ilya Sutskever. 2013{\natexlab{a}}.
\newblock Exploiting similarities among languages for machine translation.
\newblock \emph{arXiv preprint arXiv:1309.4168}.

\bibitem[{Mikolov et~al.(2013{\natexlab{b}})Mikolov, Sutskever
  et~al.}]{mikolov2013word2vec}
Tomas Mikolov, Ilya Sutskever, et~al. 2013{\natexlab{b}}.
\newblock Distributed representations of words and phrases and their
  compositionality.
\newblock In \emph{Advances in Neural Information Processing Systems}.

\bibitem[{Mnih et~al.(2016)Mnih, Badia, Mirza, Graves, Lillicrap, Harley,
  Silver, and Kavukcuoglu}]{mnih2016asynchronous}
Volodymyr Mnih, Adria~Puigdomenech Badia, Mehdi Mirza, Alex Graves, Timothy
  Lillicrap, Tim Harley, David Silver, and Koray Kavukcuoglu. 2016.
\newblock Asynchronous methods for deep reinforcement learning.
\newblock In \emph{International Conference on Machine Learning}, pages
  1928--1937.

\bibitem[{Mueller and Thyagarajan(2016)}]{mueller2016siamese}
Jonas Mueller and Aditya Thyagarajan. 2016.
\newblock Siamese recurrent architectures for learning sentence similarity.
\newblock In \emph{Proceedings of the AAAI Conference on Artificial
  Intelligence}, pages 2786--2792.

\bibitem[{Nenkova et~al.(2012)Nenkova, McKeown et~al.}]{nenkova2012survey}
Ani Nenkova, Kathleen McKeown, et~al. 2012.
\newblock A survey of text summarization techniques.
\newblock In \emph{Mining text data}, pages 43--76. Springer.

\bibitem[{Peters et~al.(2018)Peters, Neumann, Iyyer, Gardner, Clark, Lee, and
  Zettlemoyer}]{peters2018deep}
Matthew Peters, Mark Neumann, Mohit Iyyer, Matt Gardner, Christopher Clark,
  Kenton Lee, and Luke Zettlemoyer. 2018.
\newblock Deep contextualized word representations.
\newblock In \emph{Proceedings of the 2018 Conference of the North American
  Chapter of the Association for Computational Linguistics: Human Language
  Technologies, Volume 1 (Long Papers)}, volume~1, pages 2227--2237.

\bibitem[{Pires et~al.(2019)Pires, Schlinger, and
  Garrette}]{pires2019multilingual}
Telmo Pires, Eva Schlinger, and Dan Garrette. 2019.
\newblock How multilingual is multilingual bert?
\newblock In \emph{ACL}.

\bibitem[{Reddi et~al.(2018)Reddi, Kale, and Kumar}]{reddi2018convergence}
Sashank~J Reddi, Satyen Kale, and Sanjiv Kumar. 2018.
\newblock On the convergence of adam and beyond.
\newblock In \emph{International Conference on Learning Representations}.

\bibitem[{Rockt{\"a}schel et~al.(2016)Rockt{\"a}schel, Grefenstette, Hermann,
  Ko{\v{c}}isk{\`y}, and Blunsom}]{rocktaschel2015reasoning}
Tim Rockt{\"a}schel, Edward Grefenstette, Karl~Moritz Hermann, Tom{\'a}{\v{s}}
  Ko{\v{c}}isk{\`y}, and Phil Blunsom. 2016.
\newblock Reasoning about entailment with neural attention.
\newblock In \emph{International Conference on Learning Representations}.

\bibitem[{Ruder et~al.(2017)Ruder, Vuli{\'c}, and S{\o}gaard}]{ruder2017survey}
Sebastian Ruder, Ivan Vuli{\'c}, and Anders S{\o}gaard. 2017.
\newblock A survey of cross-lingual word embedding models.
\newblock \emph{Journal of Artificial Intelligence Research}.

\bibitem[{Sha et~al.(2016)Sha, Chang et~al.}]{sha2016reading}
Lei Sha, Baobao Chang, et~al. 2016.
\newblock Reading and thinking: Re-read lstm unit for textual entailment
  recognition.
\newblock In \emph{Proceedings of the International Conference on Computational
  Linguistics}.

\bibitem[{Shen et~al.(2017)Shen, Lei, Barzilay, and Jaakkola}]{shen2017style}
Tianxiao Shen, Tao Lei, Regina Barzilay, and Tommi Jaakkola. 2017.
\newblock Style transfer from non-parallel text by cross-alignment.
\newblock In \emph{Advances in Neural Information Processing Systems}, pages
  6833--6844.

\bibitem[{Speer et~al.(2017)Speer, Chin, and Havasi}]{speer2017conceptnet}
Robert Speer, Joshua Chin, and Catherine Havasi. 2017.
\newblock Conceptnet 5.5: An open multilingual graph of general knowledge.
\newblock In \emph{Thirty-First AAAI Conference on Artificial Intelligence}.

\bibitem[{Upadhyay et~al.(2018)Upadhyay, Gupta, and Roth}]{upadhyay2018joint}
Shyam Upadhyay, Nitish Gupta, and Dan Roth. 2018.
\newblock Joint multilingual supervision for cross-lingual entity linking.
\newblock In \emph{Proceedings of the 2018 Conference on Empirical Methods in
  Natural Language Processing}, pages 2486--2495.

\bibitem[{Vuli{\'c} et~al.(2016)Vuli{\'c}, Korhonen et~al.}]{vulic2016role}
Ivan Vuli{\'c}, Anna Korhonen, et~al. 2016.
\newblock On the role of seed lexicons in learning bilingual word embeddings.
\newblock In \emph{Proceedings of the 54th Annual Meeting of the Association
  for Computational Linguistics (Volume 1: Long Papers)}, volume~1, pages
  247--257.

\bibitem[{Vyas and Carpuat(2016)}]{vyas2016sparse}
Yogarshi Vyas and Marine Carpuat. 2016.
\newblock Sparse bilingual word representations for cross-lingual lexical
  entailment.
\newblock In \emph{Proceedings of the 2016 Conference of the North American
  Chapter of the Association for Computational Linguistics: Human Language
  Technologies}, pages 1187--1197.

\bibitem[{Wu et~al.(2016)Wu, Schuster, Chen, Le, Norouzi, Macherey, Krikun,
  Cao, Gao, Macherey et~al.}]{wu2016google}
Yonghui Wu, Mike Schuster, Zhifeng Chen, Quoc~V Le, Mohammad Norouzi, Wolfgang
  Macherey, Maxim Krikun, Yuan Cao, Qin Gao, Klaus Macherey, et~al. 2016.
\newblock Google's neural machine translation system: Bridging the gap between
  human and machine translation.
\newblock \emph{arXiv preprint arXiv:1609.08144}.

\bibitem[{Xing et~al.(2015)Xing, Wang, Liu, and Lin}]{xing2015normalized}
Chao Xing, Dong Wang, Chao Liu, and Yiye Lin. 2015.
\newblock Normalized word embedding and orthogonal transform for bilingual word
  translation.
\newblock In \emph{Proceedings of the 2015 Conference of the North American
  Chapter of the Association for Computational Linguistics: Human Language
  Technologies}, pages 1006--1011.

\bibitem[{Yin and Sch{\"u}tze(2015)}]{yin2015convolutional}
Wenpeng Yin and Hinrich Sch{\"u}tze. 2015.
\newblock Convolutional neural network for paraphrase identification.
\newblock In \emph{Proceedings of the 2015 Conference of the North American
  Chapter of the Association for Computational Linguistics: Human Language
  Technologies}.

\bibitem[{Yin et~al.(2016)Yin, Sch{\"u}tze et~al.}]{yin2015abcnn}
Wenpeng Yin, Hinrich Sch{\"u}tze, et~al. 2016.
\newblock Abcnn: Attention-based convolutional neural network for modeling
  sentence pairs.
\newblock \emph{Transactions of the Association for Computational Linguistics},
  4(1).

\bibitem[{Zhou et~al.(2016)Zhou, Wan, and Xiao}]{zhou2016cross}
Xinjie Zhou, Xiaojun Wan, and Jianguo Xiao. 2016.
\newblock Cross-lingual sentiment classification with bilingual document
  representation learning.
\newblock In \emph{Proceedings of the 54th Annual Meeting of the Association
  for Computational Linguistics (Volume 1: Long Papers)}, volume~1, pages
  1403--1412.

\end{thebibliography}
\bibliographystyle{acl_natbib}

\end{document}